\documentclass{article}
\usepackage{spconf,amsmath,graphicx,hyperref}
\usepackage{amsfonts}
\usepackage{multirow, booktabs}

\title{BiCLIP: Bidirectional and Consistent Language-Image Processing for Robust Medical Image Segmentation}
\name{Sivan Talaei$^{1}$, Fatemeh Daneshfar$^{1,*}$\thanks{* Corresponding author.}, Abdulhady Abas Abdullah$^{2}$, Mourad Oussalah$^{3}$}
\address{
$^{1}$Department of Computer Engineering, University of Kurdistan, Sanandaj, Kurdistan, Iran\\
$^{2}$Artificial Intelligence and Innovation Center, University of Kurdistan, Erbil, Kurdistan, Iraq\\
$^{3}$Center for Machine Vision and Signal Analysis, University of Oulu, Oulu, Finland\\
}

\begin{document}
%
\maketitle
\begin{abstract}
Medical image segmentation plays a crucial role in computer-assisted diagnosis and treatment planning. Recent vision-language approaches improve semantic understanding by incorporating textual information, but their robustness under limited annotations and degraded imaging conditions remains challenging. We propose BiCLIP, a bidirectional and consistent language-image processing framework for robust medical image segmentation. BiCLIP introduces an asymmetric fusion strategy for adaptive image-text interaction and an augmentation consistency objective that enables effective learning from augmented images while preserving stable multimodal representations. Experiments on four medical segmentation datasets demonstrate competitive performance and improved robustness under limited supervision and corrupted-data scenarios. The implementation details and source code are publicly accessible at \url{https://github.com/SeivanTalaie/BiCLIP}.

\end{abstract}
\section{Introduction}
\label{sec:intro}

\begin{figure}[t]
    \centering
    \includegraphics[width=\columnwidth]{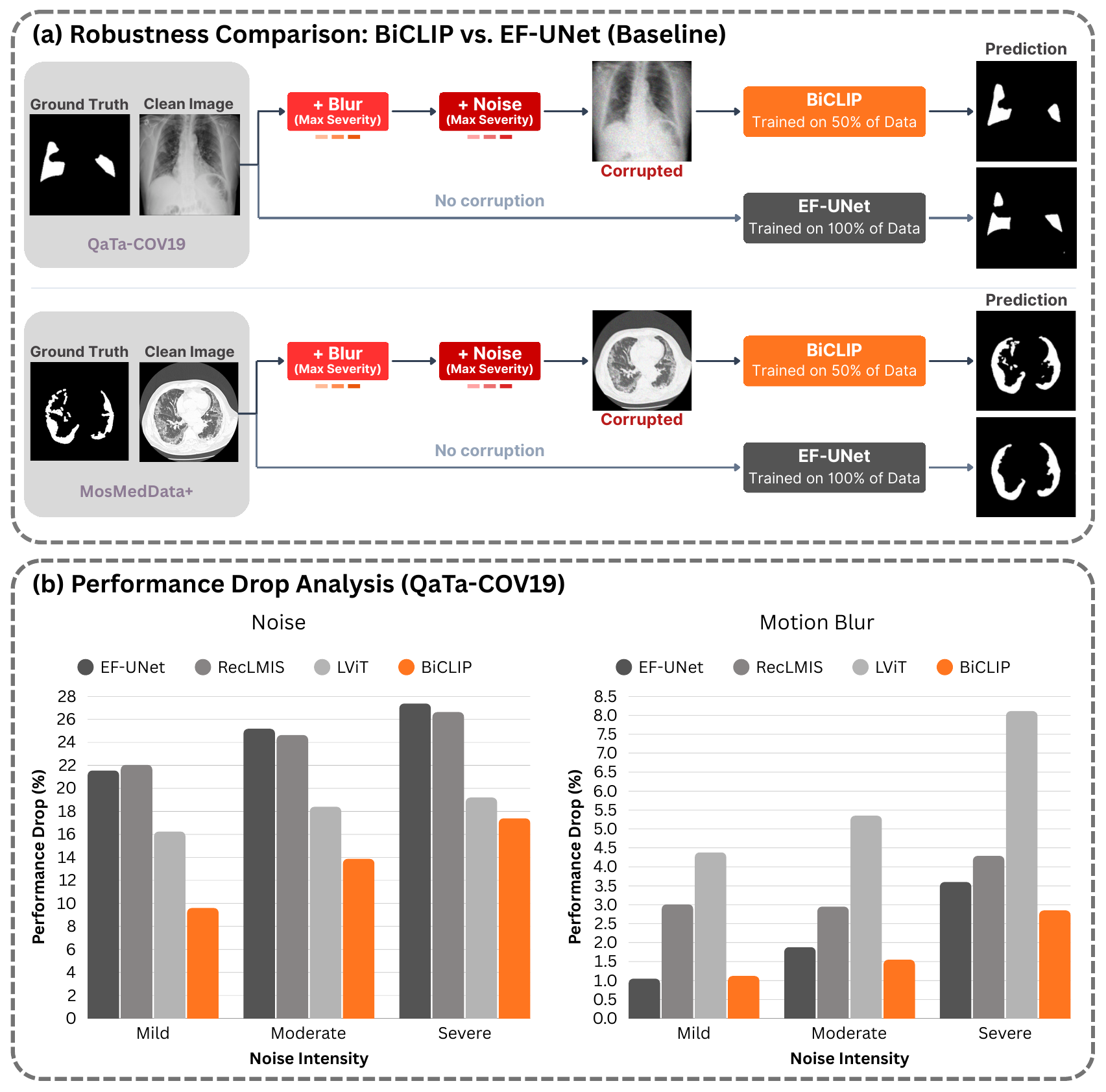}
    \caption{Robustness assessment of BiCLIP under data scarcity and image degradations. (a) Comparison with a baseline under an unfavorable training-data setting. (b) Performance degradation under different levels of noise and motion blur.}
    \label{fig1:robustness}
\end{figure} 


Deep learning has become a key approach for medical image segmentation, with models such as U-Net \cite{ronneberger2015unet} effectively learning spatial and contextual representations. However, image-based methods remain vulnerable to variations in acquisition conditions, image quality, and data distribution \cite{litjens2017survey}, especially when annotations are limited or inputs are degraded. Vision--language approaches address these challenges by incorporating semantic information from text to improve robustness and adaptability.

Integrating textual information into visual reasoning improves semantic understanding beyond appearance-based cues in segmentation. Recent approaches, such as LViT \cite{li2023lvit} and RecLMIS \cite{huang2024reclmis}, enhance segmentation through language guidance and cross-modal reconstruction, respectively. Despite these advances, maintaining effective image--text alignment remains challenging, as textual descriptions may not adequately reflect the visual characteristics of individual images. Additionally, spatial augmentations can modify the location of target regions without updating the associated descriptions, resulting in semantic inconsistencies.

\begin{figure*}[t]
    \centering
    \includegraphics[width=0.7\textwidth]{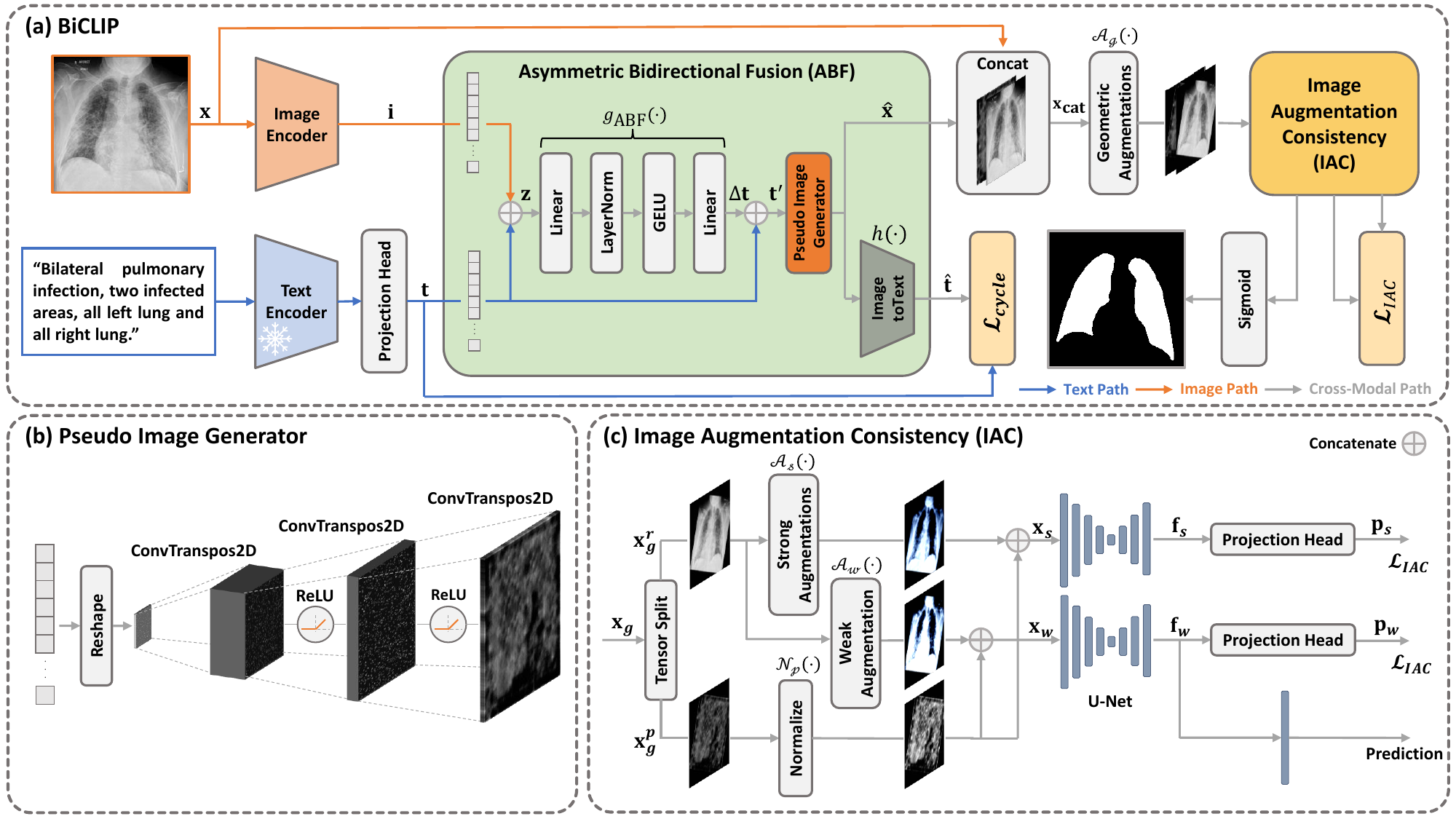}
    \caption{Overview of the BiCLIP framework. (a) Multimodal segmentation architecture, (b) pseudo-image generation for vision-language interaction, and (c) IAC module for robust learning.}
    \label{fig:architecture}
\end{figure*}

To address these limitations, we propose BiCLIP, a language-image processing framework for medical image segmentation. Its Asymmetric Bidirectional Fusion (ABF) module adapts textual representations using visual features and projects refined semantics back into image space, allowing data augmentation while preserving image--text correspondence. BiCLIP also introduces Image Augmentation Consistency (IAC), which aligns weakly and strongly augmented views to promote stable, invariant features. As shown in Fig.~\ref{fig1:robustness}, BiCLIP remains robust under limited training data and severe image degradations. The main contributions of this work are summarized as follows:





1. We propose BiCLIP, a vision--language framework using ABF for image-conditioned semantic refinement, cross-modal guidance, and reliable multimodal learning under augmentation.

2. We introduce IAC to align weakly and strongly augmented views, promoting invariant representations and improving robustness to image degradations.

3. We evaluate BiCLIP under limited annotations and corrupted images, demonstrating segmentation performance and improved robustness to data scarcity and degradation.

4. We validate BiCLIP across COVID-19 and polyp segmentation tasks, demonstrating its generalizability across diverse medical imaging scenarios, modalities, and settings.

\begin{table*}[t]

\begin{minipage}[t]{0.49\textwidth}
\centering
\caption{Quantitative segmentation results on the QaTa-COV19 and MosMedData+ datasets.}
\vspace{5pt}
\label{tab:sota}

\resizebox{\linewidth}{!}{
\begin{tabular}{lccccc}
\toprule
\multirow{2}{*}{Method} &
\multirow{2}{*}{Backbone} &
\multicolumn{2}{c}{QaTa-COV19} &
\multicolumn{2}{c}{MosMedData+} \\

\cmidrule(lr){3-4}
\cmidrule(lr){5-6}

& &
Dice(\%) & mIoU(\%) &
Dice(\%) & mIoU(\%) \\
\midrule

\multicolumn{6}{l}{\textit{Monomodal models}} \\

U-Net \cite{ronneberger2015unet}
& CNN
& 79.02
& 69.46
& 64.60
& 50.73 \\

UNet++ \cite{zhou2018unetplusplus}
& CNN
& 79.62
& 70.25
& 71.75
& 58.39 \\





\midrule

\multicolumn{6}{l}{\textit{Multimodal models}} \\



LViT \cite{li2023lvit}
& Hybrid
& 83.66
& 75.11
& 74.57
& 61.33 \\


RecLMIS \cite{huang2024reclmis}
& CNN
& 85.22
& 77.00
& 77.48
& 65.07 \\


ARSeg \cite{wang2025refseg}
& Hybrid
& 84.09
& 72.64
& 73.24
& 59.82 \\





ProLearn \cite{ye2025alleviating}
& CNN
& \textbf{90.60}
& \textbf{82.81}
& 77.82
& 63.69 \\

EF-UNet \cite{chai2025textimagefusion}
& CNN
& 90.46
& 82.58
& 80.50
& 67.37 \\


\midrule

\textbf{BiCLIP}
& CNN
& \textbf{90.60}
& \textbf{82.81}
& \textbf{81.00}
& \textbf{68.06} \\

\bottomrule
\end{tabular}
}
\end{minipage}
\hfill
\begin{minipage}[t]{0.49\textwidth}
\centering
\caption{Comparison with state-of-the-art methods on Kvasir-SEG and CVC-ClinicDB.}
\vspace{5pt}
\label{tab:polyp_sota_comparison}

\resizebox{\linewidth}{!}{
\begin{tabular}{lccccc}
\toprule
\multirow{2}{*}{Method} &
\multirow{2}{*}{Backbone} &
\multicolumn{2}{c}{Kvasir-SEG} &
\multicolumn{2}{c}{CVC-ClinicDB} \\

\cmidrule(lr){3-4}
\cmidrule(lr){5-6}

& &
Dice(\%) & mIoU(\%) &
Dice(\%) & mIoU(\%) \\
\midrule

\multicolumn{6}{l}{\textit{SOTA models}} \\

U-Net~\cite{ronneberger2015unet}
& CNN
& 81.80
& 74.60
& 82.30
& 75.50 \\


U-Net++~\cite{zhou2018unetplusplus}
& CNN
& 82.10
& 74.30
& 79.40
& 72.90 \\





LViT~\cite{li2023lvit}
& Hybrid
& 87.59
& 75.16
& 88.44
& 80.90 \\

RecLMIS~\cite{huang2024reclmis}
& CNN
& 58.08
& 44.21
& 52.21
& 38.87 \\

EF-UNet~\cite{chai2025textimagefusion}
& CNN
& 88.72
& 79.72
& 91.00
& 83.48 \\

\midrule

\multicolumn{6}{l}{\textit{SAM-based models}} \\


SAM-L~\cite{sam}
& ViT
& 78.20
& 71.00
& 57.90
& 52.60 \\

SAM-Adapter~\cite{samAdapter}
& ViT
& 84.70
& 76.30
& 77.40
& 67.30 \\




MedSAM~\cite{ma2024medsam}
& ViT
& 86.20
& 79.50
& 86.70
& 80.30 \\

\midrule

\textbf{BiCLIP}
& CNN
& \textbf{89.02}
& \textbf{80.21}
& \textbf{91.11}
& \textbf{83.68} \\

\bottomrule
\end{tabular}
}
\end{minipage}
\end{table*}

\section{Proposed Method}
\label{sec:format}

\subsection{Overview}
BiCLIP consists of parallel image and text encoders, an Asymmetric Bidirectional Fusion (ABF) module, and an Image Augmentation Consistency (IAC) module. The ABF module incorporates the pseudo-image generator for cross-modal interaction, while the IAC module contains the U-Net-based segmentation network for robust prediction. The framework is optimized using segmentation, generation, cycle-consistency, and augmentation consistency losses. Fig.~\ref{fig:architecture} illustrates the overall architecture.

\subsection{Asymmetric Bidirectional Fusion (ABF)}
Given an image $\mathbf{x}\in\mathbb{R}^{H\times W\times C}$ and its clinical description, ABF first extracts modality-specific embeddings. A frozen CXR-BERT \cite{boecking2022making} encoder produces the text representation, which is projected into $\mathbf{t}\in\mathbb{R}^{M}$. A lightweight convolutional encoder extracts the visual embedding $\mathbf{i}\in\mathbb{R}^{N}$, which is concatenated with the textual embedding $\mathbf{t}$ to form the joint representation $\mathbf{z}\in\mathbb{R}^{M+N}$.
An MLP $g_{\mathrm{ABF}}$ predicts a refinement term $\Delta\mathbf{t}$ and generates the updated text representation:
\begin{align}
\mathbf{t}'=\mathbf{t}+g_{\mathrm{ABF}}(\mathbf{z}).
\end{align}
The refined embedding is reshaped into a low-resolution spatial map $\mathbf{k}\in\mathbb{R}^{A\times A\times1}$, where $A$ denotes the spatial dimension of the intermediate feature map, and transformed by generator $G$ into a pseudo-image $\hat{\mathbf{x}}\in\mathbb{R}^{H\times W\times C}$. Finally, an image-to-text head $h$ maps the generated pseudo-image back into the text embedding space to reconstruct $\hat{\mathbf{t}}=h(\hat{\mathbf{x}})$. The reconstructed embedding is optimized using the cycle-consistency objective:
\begin{align}
\mathcal{L}_{\mathrm{cycle}}=\|\mathbf{t}-\hat{\mathbf{t}}\|_2^2.
\end{align}
This cycle constraint preserves semantic consistency across modalities by encouraging the bidirectional transformation to retain the original clinical information throughout the fusion process.

\subsection{Image Augmentation Consistency (IAC)}

After generating the pseudo-image $\hat{\mathbf{x}}$, IAC combines it with the original image $\mathbf{x}$ to construct the multimodal representation $\mathbf{x}_{\mathrm{cat}}$, which is then forwarded to the segmentation network. To improve robustness, a geometric transformation $\mathcal{A}_{g}$ is jointly applied to the multimodal input and segmentation mask $\mathbf{y}$, producing $\mathbf{x}_{g}$ and $\mathbf{y}_{g}$. The augmented input is decomposed into real and pseudo-image components, denoted as $\mathbf{x}_{g}^{r}$ and $\mathbf{x}_{g}^{p}$. Two views are then generated by applying weak and strong appearance transformations to the real-image branch while keeping the normalized pseudo-image branch unchanged:
\begin{align}
\mathbf{x}_{w}&=\mathrm{concat}(\mathcal{A}_{w}(\mathbf{x}_{g}^{r}),\mathcal{N}_{p}(\mathbf{x}_{g}^{p})),\\
\mathbf{x}_{s}&=\mathrm{concat}(\mathcal{A}_{s}(\mathbf{x}_{g}^{r}),\mathcal{N}_{p}(\mathbf{x}_{g}^{p})).
\end{align}
The shared U-Net encoder-decoder processes both views and extracts decoder features $\mathbf{f}_{w}$ and $\mathbf{f}_{s}$. A projection head $P$ maps these features into compact representations $\mathbf{p}_{w}$ and $\mathbf{p}_{s}$. The IAC loss enforces consistency between the two representations:
\begin{align}
\mathcal{L}_{\mathrm{IAC}}=1-\frac{\mathbf{p}_{w}^{\top}\mathbf{p}_{s}}
{\|\mathbf{p}_{w}\|_{2}\|\mathbf{p}_{s}\|_{2}}.
\end{align}
By encouraging invariance to appearance variations, IAC improves the stability of learned segmentation features while preserving the semantic guidance provided by the pseudo-image.

\subsection{Overall Loss Function}

The training objective of BiCLIP jointly optimizes segmentation accuracy, pseudo-image quality, multimodal semantic alignment, and feature robustness. The total loss combines the segmentation loss $\mathcal{L}_{\mathrm{seg}}$, generation loss $\mathcal{L}_{\mathrm{gen}}$, augmentation consistency loss $\mathcal{L}_{\mathrm{IAC}}$, and cycle-consistency loss $\mathcal{L}_{\mathrm{cycle}}$:
\begin{equation}
\mathcal{L}_{\mathrm{total}}=
\mathcal{L}_{\mathrm{seg}}+
\lambda_{\mathrm{gen}}\mathcal{L}_{\mathrm{gen}}+
\lambda_{\mathrm{IAC}}\mathcal{L}_{\mathrm{IAC}}+
\lambda_{\mathrm{cycle}}\mathcal{L}_{\mathrm{cycle}}.
\end{equation}
Here, $\mathcal{L}_{\mathrm{seg}}$ uses Dice and cross-entropy supervision between predicted and ground-truth masks, while $\mathcal{L}_{\mathrm{gen}}$ applies an $L_1$ reconstruction loss between generated pseudo-images and masked image supervision. The auxiliary loss weights were set to $\lambda_{\mathrm{gen}}=0.1$, $\lambda_{\mathrm{IAC}}=0.05$, and $\lambda_{\mathrm{cycle}}=0.07$.

\section{Experiments and Results}

\subsection{Datasets}

Experiments were conducted on four public datasets: QaTa-COV19 \cite{Qata1}, MosMedData+ \cite{morozov2020mosmeddata}, Kvasir-SEG \cite{jha2019kvasir}, and CVC-ClinicDB \cite{BERNAL201599}. Following the LViT \cite{li2023lvit} protocol, we use the same dataset splits and adopt text annotations from \cite{li2023lvit,poudel2023exploring}. The dataset splits are:\\
QaTa-COV19: Training=5716; Validation=1429; Test=2113\\
MosMedData+: Training=2183; Validation=273; Test=273 \\
Kvasir-SEG: Training=800; Validation=100; Test=100 \\
CVC-ClinicDB: Training=490; Validation=61; Test=61.

\subsection{Implementation Details}
All experiments were implemented in PyTorch and conducted on a single NVIDIA RTX 4090 GPU. Models were optimized using AdamW with an initial learning rate of $1\times10^{-4}$ and a cosine annealing warm restart scheduler. All methods were trained for 150 epochs with a batch size of 16 under the same settings. Input images and masks were resized to $224\times224$ for consistent evaluation.

\begin{table*}[t]
	\centering
	\footnotesize
	\caption{Quantitative robustness analysis on corrupted test images. Models are evaluated under three levels of additive Poisson noise and three motion-blur kernel sizes on QaTa-COV19 and MosMedData+.}
    \vspace{3pt}
	\label{tab:noise_robustness}
    \renewcommand{\arraystretch}{0.86}
	\resizebox{\linewidth}{!}{
	\begin{tabular}{lcccccccccc}
		\toprule
		\multirow{3}{*}{Method} &
		\multicolumn{5}{c}{\textbf{Noise Corruption}} &
		\multicolumn{5}{c}{\textbf{Motion Blur}} \\
		\cmidrule(lr){2-6}\cmidrule(lr){7-11}
		& \multirow{2}{*}{Noise} &
		\multicolumn{2}{c}{QaTa-COV19} &
		\multicolumn{2}{c}{MosMedData+} &
		\multirow{2}{*}{Kernel} &
		\multicolumn{2}{c}{QaTa-COV19} &
		\multicolumn{2}{c}{MosMedData+} \\
		\cmidrule(lr){3-4}\cmidrule(lr){5-6}\cmidrule(lr){8-9}\cmidrule(lr){10-11}
		& & Dice(\%) & mIoU(\%) & Dice(\%) & mIoU(\%) & & Dice(\%) & mIoU(\%) & Dice(\%) & mIoU(\%) \\
		\midrule
		
		\multirow{3}{*}{LViT \cite{li2023lvit}}
		& 140 & 70.07 & 58.65 & 54.57 & 43.75
		& K3  & 79.99 & 70.85 & 59.03 & 47.71 \\
		& 120 & 68.27 & 56.71 & 54.19 & 43.28
		& K5  & 79.18 & 69.88 & 56.18 & 44.50 \\
		& 110 & 67.60 & 55.95 & 52.68 & 41.95
		& K7  & 76.87 & 67.20 & 50.93 & 39.14 \\
		\midrule
		
		\multirow{3}{*}{RecLMIS \cite{huang2024reclmis}}
		& 140 & 66.44 & 53.89 & 64.16 & 52.52
		& K3  & 82.65 & 73.80 & 76.77 & 64.15 \\
		& 120 & 64.23 & 51.52 & 63.36 & 51.87
		& K5  & 82.70 & 73.79 & 75.60 & 62.61 \\
		& 110 & 62.53 & 49.77 & 62.79 & 51.37
		& K7  & 81.56 & 72.29 & 71.86 & 58.21 \\
		\midrule
		
		\multirow{3}{*}{EF-UNet \cite{chai2025textimagefusion}}
		& 140 & 70.97 & 55.00 & 79.42 & 65.87
		& K3  & 89.51 & 81.01 & 80.49 & 67.35 \\
		& 120 & 67.68 & 51.15 & 79.26 & 65.65
		& K5  & 88.76 & 79.80 & 80.15 & 66.88 \\
		& 110 & 65.70 & 48.92 & 79.03 & 65.33
		& K7  & 87.20 & 77.31 & 78.76 & 64.97 \\
		\midrule
		
		\multirow{3}{*}{\textbf{BiCLIP}}
		& 140 & \textbf{81.90} & \textbf{69.35} & \textbf{79.88} & \textbf{66.51}
		& K3  & \textbf{89.58} & \textbf{81.13} & \textbf{81.00} & \textbf{68.07} \\
		& 120 & \textbf{78.03} & \textbf{63.97} & \textbf{79.64} & \textbf{66.17}
		& K5  & \textbf{89.19} & \textbf{80.50} & \textbf{80.66} & \textbf{67.59} \\
		& 110 & \textbf{74.84} & \textbf{59.79} & \textbf{79.60} & \textbf{66.11}
		& K7  & \textbf{88.01} & \textbf{78.59} & \textbf{79.27} & \textbf{65.66} \\
		\bottomrule
	\end{tabular}}
	
\end{table*}

\begin{figure}[t]
    \centering
    \includegraphics[width=\columnwidth]{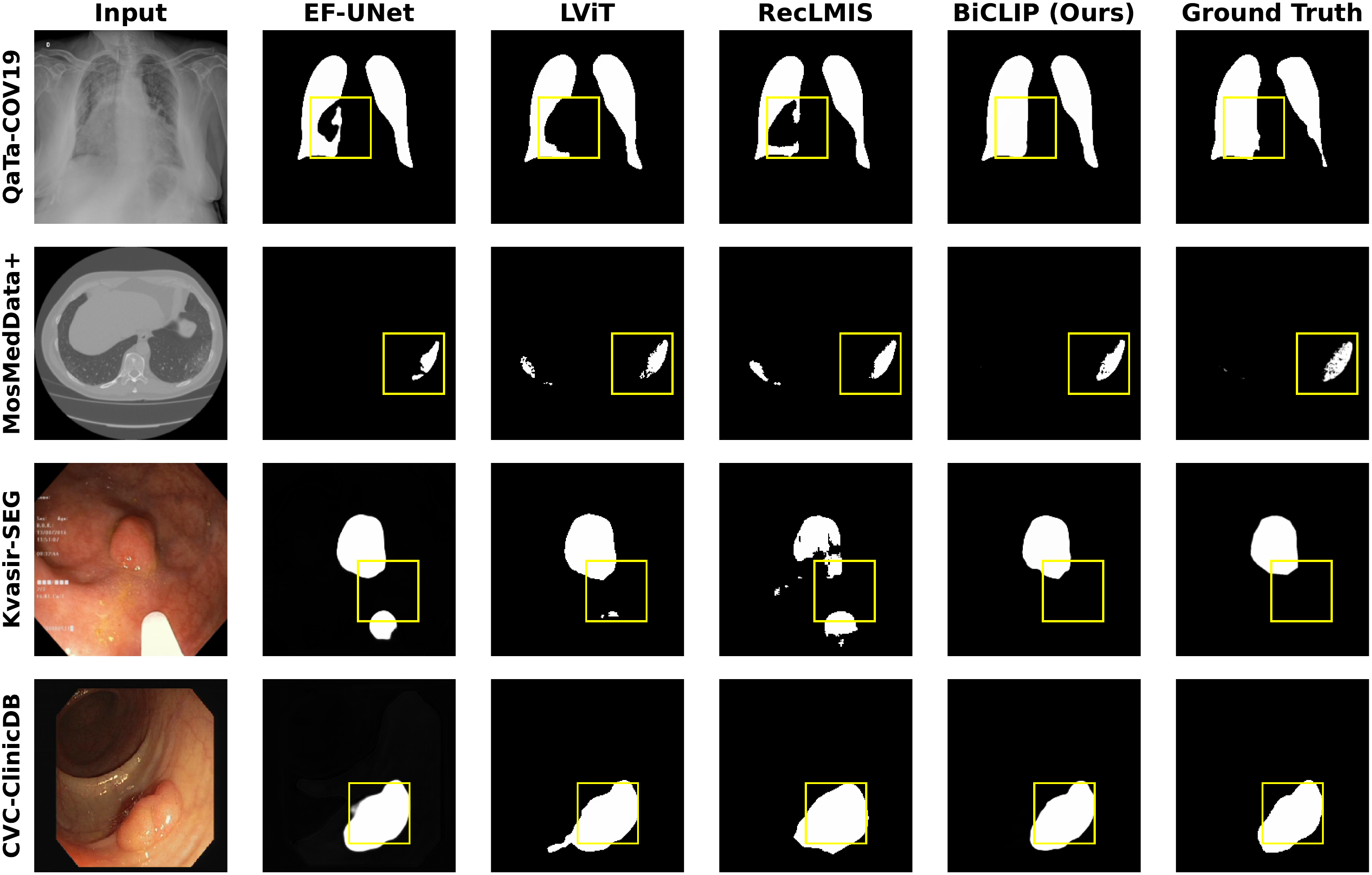}
    \caption{Qualitative segmentation results across four datasets.}
    \label{fig3:qualitative}
\end{figure} 

\begin{table}[!t]
\centering
\caption{Comparison of MLP and cross-attention fusion in ABF on QaTa-COV19 and Kvasir-SEG.}
\vspace{5pt}
\label{tab:abf_fusion_ablation}
\resizebox{\linewidth}{!}{
\begin{tabular}{lcccc}
\toprule
\multirow{2}{*}{ABF Fusion Strategy} &
\multicolumn{2}{c}{QaTa-COV19} &
\multicolumn{2}{c}{Kvasir-SEG} \\
\cmidrule(lr){2-3}
\cmidrule(lr){4-5}
& Dice(\%)  & mIoU(\%) 
& Dice(\%)  & mIoU(\%)  \\
\midrule
Cross-attention ABF
& 90.55 & 82.74
& 87.99 & 78.55 \\

MLP-based ABF
& \textbf{90.60} & \textbf{82.81}
& \textbf{89.02} & \textbf{80.21} \\
\bottomrule
\end{tabular}}
\end{table}

\subsection{Experimental Results}

Tables~\ref{tab:sota} and~\ref{tab:polyp_sota_comparison} compare BiCLIP with representative methods across COVID-19 and polyp datasets. BiCLIP achieves competitive Dice and mIoU performance while requiring 47.20 G FLOPs and reaching 148.102 images/s throughput, demonstrating a favorable balance between accuracy and efficiency. Qualitative comparisons in Fig.~\ref{fig3:qualitative} further illustrate that BiCLIP produces accurate localization and boundary delineation, with predictions closely matching the ground-truth structures across multiple datasets. Robustness evaluations in Tables~\ref{tab:noise_robustness} and~\ref{tab:robustness_unified} assess the model under limited data and degraded input conditions. The corrupted images are generated using Poisson noise with different intensity levels (where 110 indicates stronger degradation than 140) and directional motion blur kernels (where K7 represents more severe blur than K3). Beyond these evaluations, we further analyze the effect of ABF fusion strategies in Table~\ref{tab:abf_fusion_ablation}, where the MLP-based design provides better performance than cross-attention fusion. We also examine run-to-run stability using five independent training runs with seeds starting from 44, obtaining Dice standard deviations of 0.09\% and 0.52\% on QaTa-COV19 and Kvasir-SEG, respectively. Finally, the ablation study in Table~\ref{tab:ablation_losses} confirms the complementary contributions of IAC and Cycle losses, with their combination achieving the best results.





\begin{table}[!t]
	\centering
    \footnotesize
	\caption{Robustness evaluation across progressively smaller fractions of available training data.}
    \vspace{5pt}
	\label{tab:robustness_unified}
    \resizebox{\linewidth}{!}{
	\begin{tabular}{llcccc}
		\toprule
		\multirow{2}{*}{Method} &
		\multirow{2}{*}{Regime} &
		\multicolumn{2}{c}{QaTa-COV19} &
		\multicolumn{2}{c}{MosMedData+} \\
		\cmidrule(lr){3-4} \cmidrule(lr){5-6} 
		& & Dice(\%) & mIoU(\%) & Dice(\%) & mIoU(\%) \\ 
		\midrule
		\multirow{5}{*}{EF-UNet~\cite{chai2025textimagefusion}}
		& 50\% & \textbf{89.85} & \textbf{81.57} & 74.89 & 59.86 \\ 
		& 25\% & 88.78 & 79.82 & 65.63 & 48.85 \\ 
		& 10\% & \textbf{87.84} & \textbf{78.32} & 64.24 & 47.32 \\ 
		& 5\%  & 84.87 & 73.71 & 55.48 & 38.39 \\ 
		& 1\%  & 66.76 & 50.10 & 33.68 & 20.25 \\ 
		\midrule
		\multirow{5}{*}{\textbf{BiCLIP}}
		& 50\% & 89.83 & 81.53 & \textbf{79.72} & \textbf{66.27} \\ 
		& 25\% & \textbf{88.91} & \textbf{80.04} & \textbf{78.11} & \textbf{64.09} \\ 
		& 10\% & 87.16 & 77.25 & \textbf{69.23} & \textbf{52.95} \\ 
		& 5\%  & \textbf{85.13} & \textbf{74.11} & \textbf{64.71} & \textbf{47.83} \\ 
		& 1\%  & \textbf{76.63} & \textbf{62.12} & \textbf{50.42} & \textbf{33.71} \\ 
		\bottomrule
	\end{tabular}}
\end{table}



\begin{table}[!t]
\centering
\footnotesize
\caption{Ablation study of $\mathcal{L}_{\text{IAC}}$ and $\mathcal{L}_{\text{Cycle}}$ across datasets.}
\vspace{5pt}
\label{tab:ablation_losses}
\resizebox{\linewidth}{!}{
\begin{tabular}{cccccc}
\hline
\multirow{2}{*}{$\mathcal{L}_{\text{IAC}}$} &
\multirow{2}{*}{$\mathcal{L}_{\text{Cycle}}$} &
\multicolumn{2}{c}{QaTa-COV19} &
\multicolumn{2}{c}{MosMedData+} \\
\cmidrule(lr){3-4} \cmidrule(lr){5-6}
& & Dice(\%) & mIoU(\%) & Dice(\%) & mIoU(\%) \\
\hline
$\times$ & $\times$ & 90.46 & 82.58 & 80.50 & 67.37 \\
$\checkmark$ & $\times$ & 90.52 & 82.69 & 80.52 & 67.39 \\
$\times$ & $\checkmark$ & 90.57 & 82.77 & 80.53 & 67.42 \\
$\checkmark$ & $\checkmark$ & \textbf{90.60} & \textbf{82.81} & \textbf{81.00} & \textbf{68.06} \\
\hline
\end{tabular}}
\end{table}

\section{Conclusion}



This work introduced BiCLIP, a vision--language framework for medical image segmentation with adaptive fusion and consistency learning. Experiments demonstrate competitive accuracy, efficiency, robustness to degradation, and stable performance across diverse medical datasets.

\bibliographystyle{IEEEbib}
\bibliography{strings,refs}

\end{document}